\documentclass[journal,twoside,web]{ieeecolor}
\pdfoutput=1
\usepackage{generic}
\usepackage{cite}
\usepackage{amsmath,amssymb,amsfonts}
\usepackage{algorithmic}
\usepackage{graphicx}
\usepackage{textcomp}
\usepackage{makecell}
\def\BibTeX{{\rm B\kern-.05em{\sc i\kern-.025em b}\kern-.08em
    T\kern-.1667em\lower.7ex\hbox{E}\kern-.125emX}}
\begin{document}
\title{SHAPE: A Sample-adaptive Hierarchical Prediction Network for Medication Recommendation }
\author{Sicen Liu, Xiaolong Wang, Jingcheng Du, Yongshuai Hou, Xianbing Zhao, Hui Xu, Hui Wang, Yang Xiang, \IEEEmembership{Member, IEEE}, Buzhou Tang, \IEEEmembership{Member, IEEE}
\thanks{ This study is partially supported by the National Natural Science Foundations of China (62276075, 62276082, U1813215, 61876052 and 62106115), the Science and Technology Planning Project of Shenzhen Municipality (JCYJ20190806112210067), the National Key R\&D Program of China (2021ZD0113402), the National Natural Science Foundation of Guangdong, China (2019A1515011158), Major Key Project of PCL (PCL2021A06), Strategic Emerging Industry Development Special Fund of Shenzhen (20200821174109001). (Corresponding author: Yang Xiang, Buzhou Tang).}
\thanks{Sicen Liu, Xianbing Zhao and Buzhou Tang are now with the Department of Computer Science, Harbin Institute of Technology (Shenzhen), Shenzhen, China, and Peng Cheng Laboratory, Shenzhen, China (email: liusicen@stu.hit.edu.cn; tangbuzhou@gmail.com, zhaoxianbing\_hitsz@163.com).}
\thanks{Jingcheng Du is now with Melax Tech, Inc., Houston, Texas, United States, and The University of Texas Health Science Center at Houston, Texas, United States (email: jingcheng.du@melaxtech.com)}
\thanks{Xiaolong Wang is now with the Department of Computer Science, Harbin Institute of Technology (Shenzhen), Shenzhen, China (email: wangxl@insun.hit.edu.cn.) }
\thanks{Hui Xu and Hui Wang are now with Gennlife(Beijing) Technology Co Ltd, Beijing, China (email: wanghui@gennlife.com;xuhui@gennlife.com)}
\thanks{Yongshuai Hou and Yang Xiang are now with Peng Cheng Laboratory, Shenzhen, China (email: houysh@pcl.ac.cn; xiangy@pcl.ac.cn).}}

\maketitle

\begin{abstract}
Effectively medication recommendation with complex multimorbidity conditions is a critical task in healthcare. Most existing works predicted medications based on longitudinal records, which assumed the information transmitted patterns of learning longitudinal sequence data are stable and intra-visit medical events are serialized. However, the following conditions may have been ignored: 1) A more compact encoder for intra-relationship in the intra-visit medical event is urgent; 2) Strategies for learning accurate representations of the variable longitudinal sequences of patients are different. In this paper, we proposed a novel Sample-adaptive Hierarchical medicAtion Prediction nEtwork, termed SHAPE, to tackle the above challenges in the medication recommendation task. Specifically, we design a compact intra-visit set encoder to encode the relationship in the medical event for obtaining visit-level representation and then develop an inter-visit longitudinal encoder to learn the patient-level longitudinal representation efficiently. To endow the model with the capability of modeling the variable visit length, we introduce a soft curriculum learning method to assign the difficulty of each sample automatically by the visit length. 
Extensive experiments on a benchmark dataset verify the superiority of our model compared with several state-of-the-art baselines.

\end{abstract}


\section{Introduction}
\label{sec:introduction}

\IEEEPARstart{R}{ecently}, massive health data have offered the opportunity to assist clinical decision-making through deep learning \cite{b1,b2,b3,b4,liu2020hybrid,MCF}. Effective and safe medication combination recommendation for patients who suffer from multiple diseases is an essential task in healthcare \cite{b5,symeonidis2021recommending,an2021mesin}. 
There are a lot of research interests on medication recommendation task \cite{b6,G-Bert,b7,b8, b9,b10, ARMR, wang2021self, wang2021multi, si2021deep}. The intuitive goal of medication recommendation is to predict medication sequences for a particular patient based on complex health conditions.  Existing strategies of medication recommendation can be categorized into two types: 1) \textit{Instance-based  methods}, which recommend medication sequences only based on the current hospital visit(e.g., diagnosis,  procedure) \cite{smr,safeMedicinRs, leap, 4sdrug}. The instance-based setting will ignore the temporal dependencies on the patient's health records. To overcome this issue, 2) \textit{Longitudinal-based methods} were proposed to leverage the longitudinal patient records to predict personalized medication. Most longitudinal methods pursue enhanced representations of patient health status based on the historical health records(e.g., diagnosis, procedure) and use this patient representation to conduct medication recommendations \cite{retain,dmnc,gamenet, MICRON, safedrug, COGNet, b11, DRMP}. 

Despite the significance and value of the methods in the longitudinal methods, they still suffer from two critical limitations: \textit{1)} One problem with existing longitudinal works is that they neglect the compact intra-relationships between medical events within each visit. In other words, they ignore the relationship between the same type of medical codes during a visit. \textit{2)} Existing longitudinal models are static. Namely, all samples go through the same fixed computation flow. This may be powerless on the shorter records, which lack historical information. 

\begin{figure}[htbp]
	\centering
	\includegraphics[width=85mm]{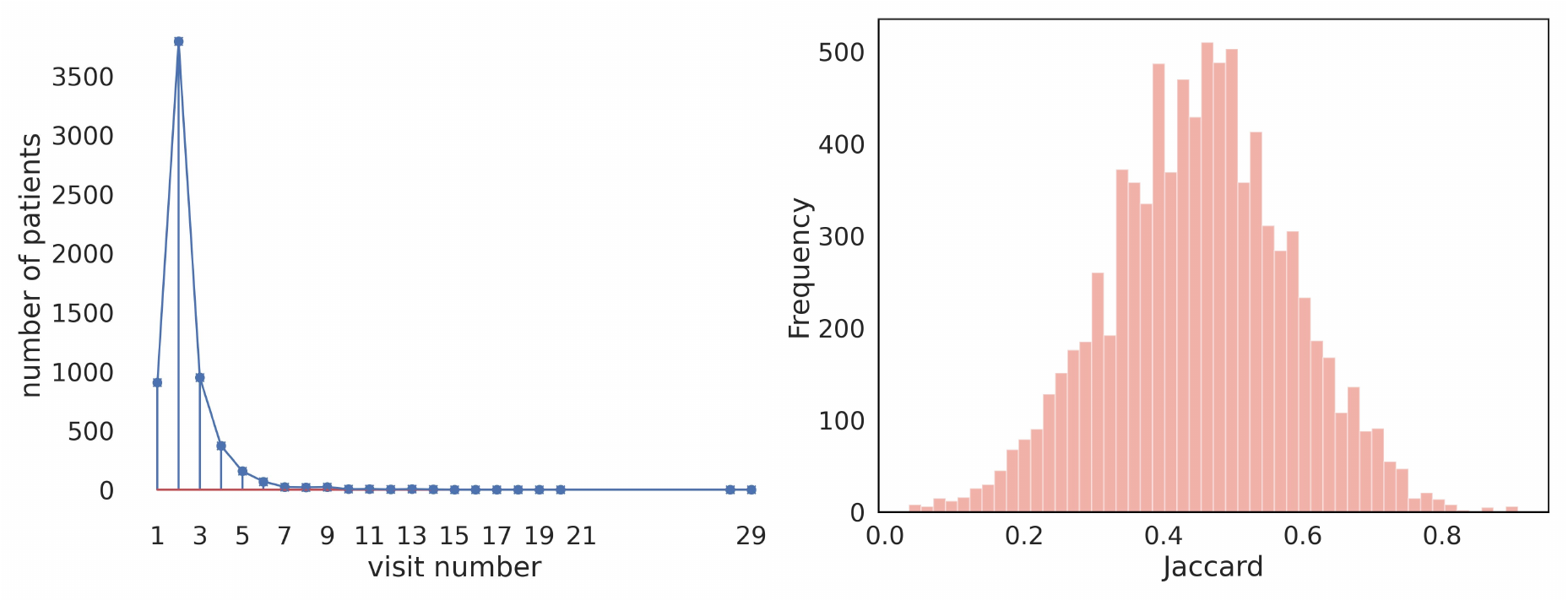}
	\caption{The histogram of visit counts of MIMIC-III dataset (left) and the histogram of Jaccard between current medications and historical medications (right).}
\end{figure}

\begin{figure}[htbp]
	\centering
	\includegraphics[width=85mm]{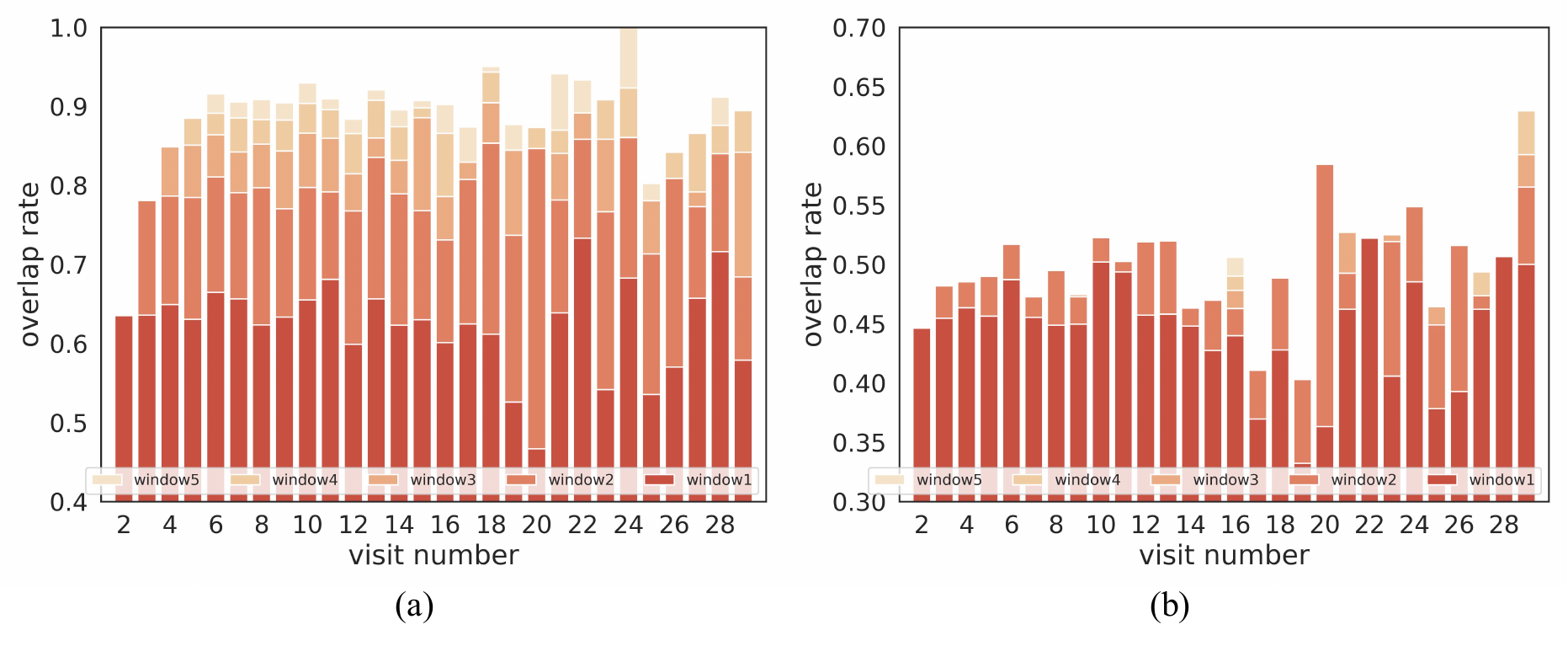}
	\caption{The statistics of (a) medication overlap rate and (b) Jaccard coefficients in various visits with different window sizes.}
\end{figure}

On the one hand, existing longitudinal methods use the historical code sequences (e.g., medication, diagnosis) within each visit to present the complex patient's health condition, where medical events are adopted independently and sparsely represented methods to obtain equal contributions representation in the current record. Most of them use multi-hot embedding methods to encode the structured data sequences. However, the impact of medical events varies for each patient, especially for patients with multimorbidity. For instance, during a visit, the health condition differs a lot between a patient diagnosed with both \textit{Chronic systolic heart failure} and \textit{Septic shock} and a patient diagnosed with both \textit{Septic shock} and \textit{Acute respiratory failure}. Previous methods ignore the compact intra-relationship of these medical events and the variable importance of each code for the patient.

On the other hand, such longitudinal patterns rely on historical health information and are powerless to the short visit that lacks historical records. As shown in Figure 1, we conduct the statistic on the MIMIC-III\cite{MIMIC-III} dataset. We can see that most visit lengths are short than thrice. For each visit, we calculate the Jaccard between current medications and past medications. We can see that a large portion of prescribed medicines are similar to those recommended before, which means the results of medication recommendations rely on historical medication records. Additionally, we conduct fine-grained statistics of the MIMIC-III dataset, as shown in Figure 2. We calculate the proportion of medications that have appeared in history and the Jaccard with various visit windows. We can see that in the more extended visits, a large portion of drug sequences have been recommended before. However, the prevalence of short visit records in real-world clinical scenarios often lacks crucial historical medication information that could be referenced for treatment decisions. 
This phenomenon illustrates that a more robust strategy that could model the accurate representation of the variable longitudinal sequences is urgent.

To overcome these challenges, we proposed a novel \textbf{S}ample-adaptive \textbf{H}ierarchical medic\textbf{A}tion \textbf{P}rediction n\textbf{E}twork, named \textbf{SHAPE}, to learn a more accurate representation of patients. In SHAPE, we develop a hierarchical patient representation framework. Concretely,
we first tailor an intra-visit set encoder to learn the visit-level representation and then design an inter-visit longitudinal encoder for learning the patient-level longitudinal representation. By performing the intra-visit set encoder and inter-visit longitudinal encoder, collaborative information latent in longitudinal historical interactions is explicitly hierarchical encoded. To enhance the ability to represent various lengths of visit records, we adopt a soft curriculum learning method to help our SHAPE model learn these data patterns by assigning the difficulty weight to each sample.
The experiments on a public dataset demonstrate the effectiveness of our proposed model. 

The main contributions of this work are three-fold:

\begin{itemize}
\item We present a hierarchical encoder mechanism towards medication recommendation, which could dig for a more accurate representation from the various records of the patient. In particular, we first design an intra-visit set encoder to encode the medical events and obtain visit-level representation, and then develop an inter-visit longitudinal encoder for learning the patient-level longitudinal information.


\item We design an adaptive curriculum learning module for variable patient visit records, especially for the short ones, which aims at an adaptive learning strategy over time and the length of patient records to improve the effectiveness of medication recommendations.
\item  Extensive experimental results on the public benchmark dataset validate the effectiveness and superiority of our proposed method.

\end{itemize}

\section{Related work}

\subsection{Medication recommendation}

Existing medication recommendation algorithms can be categorized into instance-based methods and longitudinal approaches. Instance-based algorithms extract patient information only from current visits. For example, LEAP \cite{leap} extracts patient representation from the current visit record and decomposes the medication recommendation into a sequential decision-making process. Longitudinal-based methods are designed to leverage temporal dependencies within the patient's historical information. For example, RETAIN \cite{retain} uses two-level attention, which models the longitudinal information based on recurrent neural networks (RNN). GAMENet \cite{gamenet} uses augmented memory neural networks to fuse the drug-drug interactions and store the historical drug record to model the patient representation. MICRON\cite{MICRON} pays attention to the changes in patient health records and uses residual-based network inference to update the sequential representation. COGNet \cite{COGNet} conditional generates the medication combinations either copied from the historical drug records or direct generate new drugs. 
These existing efforts, however, still suffer from the following limitations. Existing work ignores that the intra-visit medical events may pay variable effects on differing the health status of the patient. Most of them use multi-hot embedding to encode the medical events in the current visit and ignore the difference of each medical event in intra-visit records. In this paper, we proposed a hierarchical architecture to learn the comprehensive patient representation. We use an intra-visit set encoder to learn a more accurate representation of intra-visit medical events and develop an inter-visit longitudinal encoder to learn longitudinal information about the patient.  

\subsection{Curriculum learning}
The conventional curriculum learning methods formalized the organized learning process of humans and animals, which illustrates gradually more complex ones~\cite{bengio2009curriculum}. Alex et al. derived two distinct indicators (i.e., rate of increase in prediction accuracy and rate of increase in network complexity) of the learning process as the reward signal to maximize learning efficiency automatically~\cite{graves2017automated}. Guy et al. introduce sorted samples with different scoring functions to assign the learning difficulty of each instance~\cite{hacohen2019power}.  Recently, curriculum learning has been applied to different medical tasks. Basu et al. propose a curriculum inspired by human visual acuity, which reduces the texture biases for gallbladder cancer detection~\cite{basu2022surpassing}. Guo et al. demonstrate the application of curriculum learning for drug molecular design~\cite{guo2022improving}. Gu et al. utilized curriculum learning to improve the training efficiency of molecular graph learning ~\cite{gu2022efficient}.   According to Figure 1 and Figure 2, we found that the short and new visits samples account for most of the entire dataset. The conventional longitudinal methods are hard to fit this pattern because lacking a flexible ability to model the scenarios where the patients do not have enough historical medication records and diagnosis information about their health condition. In this paper, we propose a sample-adapting curriculum learning algorithm to assign the difficulty of each instance automatically. 

\section{Problem Formulation}

\subsection{Electrical Health Records (EHR)}

Patient EHR data contains comprehensive medical information about the patient. Formally, EHR for patient $j$ can be represented as a sequence $X_j = (x_j^{1}, x_j^{2}, \cdots, x_j^{T})$, where $T$ is the corresponding totally visits number for patient $j$. For the single visit $x_j^{t}$ of patient $j$ at $t-$th visit, where $t \in \{ 1,2, \cdots, T\}$, we ignore the index $j$ of patient to simplify notation. Then, the visit record is represented as $x^{t} = (D^{t}, P^{t}, M^{t})$, where $D^{t} \subseteq \{d_1, d_2, \cdots, d_{|\mathcal{D}|}\}$ denotes the set of diagnoses appeared in $t$-th visit, $P^{t} \subseteq \{ p_1, p_2, \cdots, p_{|\mathcal{P}|} \}$ denotes the set of procedures and $M^{t} \subseteq \{ m_1, m_2, \cdots, m_{|\mathcal{M}|} \}$ denotes the set of medications appeared in $t$-th visit. $|\mathcal{D}|, |\mathcal{P}|$ and $|\mathcal{M}|$ indicate the cardinality of corresponding element sets.

\subsection{DDI Graph}
The medications may interact with other medications when prescribed, while the adverse drug-drug interactions (DDIs) graph records this interaction of adverse drug events. The DDI graph can be denoted as $\mathcal{G}_d=\{ \mathcal{V}, \mathcal{E}_d \}$, where node set $\mathcal{V} \in \{ m_1, m_2, \cdots, m_{|\mathcal{M}|} \}$ represent the set of medications. The $\mathcal{E}_d$ is the edge set of known DDIs between a pair of drugs. Adjacency matrix $A_d \in \mathbb{R}^{|\mathcal{M}| \times |\mathcal{M}|}$ are defined to the construction of the graphs. When the $A_d[i,j]=1$ means the $i$-th medication and $j$-th one could interact with each other.

\subsection{Medication Recommendation Problem}
Given a patient EHR sequence $[x^1, x^2, \cdots, x^t]$ and the DDI graph $\mathcal{G}_d$. For the multi-visit records patient, which includes the current diagnosis, procedure codes $[D^{t}, P^{t}]$ and the historical records $[x^1, x^2, \cdots, x^{t-1}]$.  Note that, for the record of new visit patients, there are only current diagnosis and procedure codes $[D^{1}, P^{1}]$. The goal is to train a model to effectively recommend multiple medications by generating multi-label output $\hat{y}_t \subseteq \{ m_1, m_2, \cdots, m_{|\mathcal{M}|} \}$ for this patient.
\begin{figure*}[htbp]
	\centering
	\includegraphics[width=180mm]{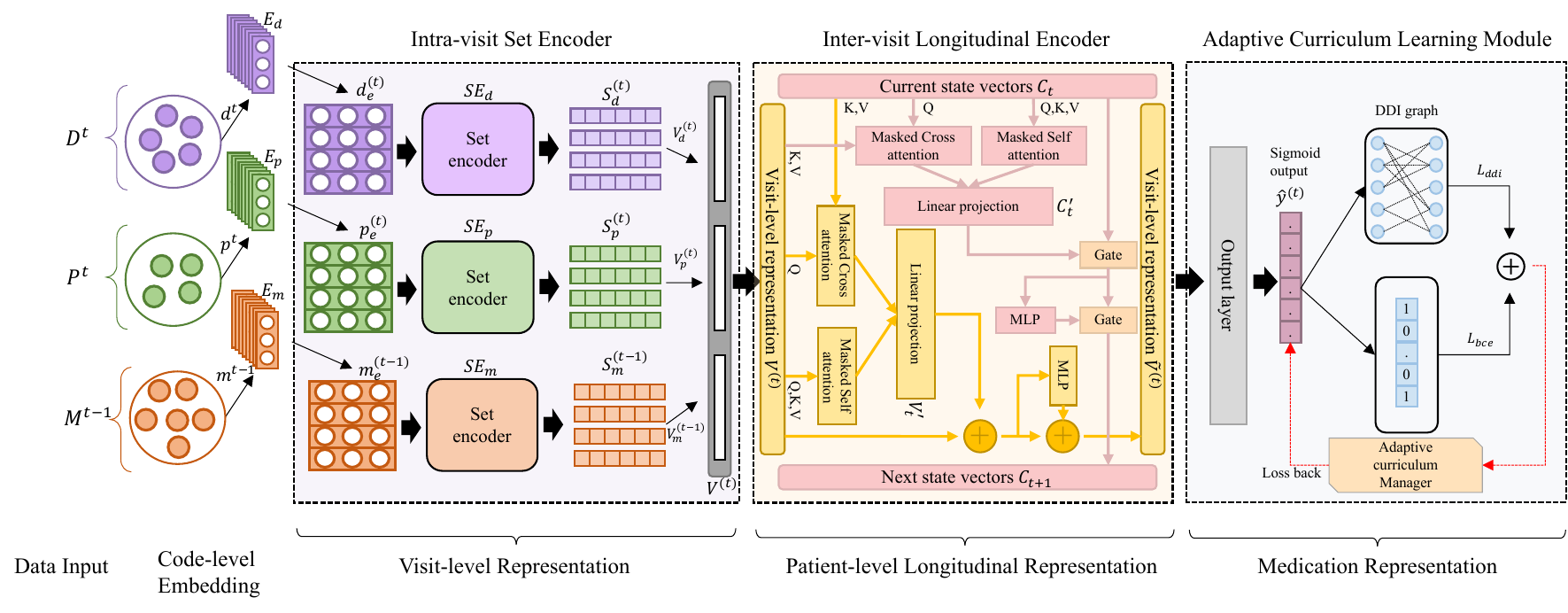}
	\caption{The Framework of our proposed SHAPE. There are three components:(1) The Intra-visit Set Encoder captures the intra-relationship of the code-level medical events and summarizes it to the current visit-level representation. (2) An Inter-visit Longitudinal Encoder to model the longitudinal information of the patient. (3) An Adaptive curriculum learning module automatically assigns each sample's difficulty according to the patient's visit length.}
\end{figure*}
\section{The SHAPE Framework}
In this section,  we present the technical details of the proposed \textbf{SHAPE} framework. As illustrated in Figure 3, our model includes three components: (1) an \textbf{intra-visit set encoder} that learns the visit-level representation of the patient from the EHR data. (2) an \textbf{inter-visit longitudinal encoder} that takes the visit-level representation as input to learn the longitudinal information of the patient. (3) a \textbf{adaptive curriculum learning module} that cooperates with the prediction phase in the training stage to dynamically assign the difficulty weight of each instance by the patient visit length to improve the effectiveness of medication recommendations. Finally, the drug output is obtained from the sigmoid output representation.

\subsection{Patient Representation}
The patient representation aims to learn a dense vector to represent a comprehensive patient's status. The physicians recommend medications based on the current diagnosis and procedure information during a clinical visit. Furthermore, the clinician also references the history of diagnosis, procedure, and medication records when the patient has historical visit records. Since the SHAPE is proposed for the generic patient,  we use the three codes as the model input in the following, and the medication codes are always behind the other two medical events. Note that, for the patient who only once visit with diagnosis and procedure record, we apply a padding embedding as the medic input.

\subsubsection{code-level embedding} For predict the medication of multi-visit, we use the $[D^{t}, P^{t}, M^{t-1}]$ as the current input, where $M^{t-1}$ is the previous medication record. We design three correspond embedding table $E_d \in \mathbb{R}^{|\mathcal{D}| \times dim}, E_p \in \mathbb{R}^{|\mathcal{P}| \times dim}$ and $ E_m \in \mathbb{R}^{|\mathcal{M}| \times dim}$, where the $dim$ is the dimension of the embedding space. For the $t-$th visit, the set of medical events $ d^{(t)}\in D^{t}, p^{(t)}\in P^{t}$, and $m^{(t-1)} \in M^{t-1}$ was transfer to the embedding space.
\begin{equation}
d_e^{(t)} = d^{(t)} E_d 
\end{equation}
\begin{equation}
p_e^{(t)} = p^{(t)} E_p 
\end{equation}
\begin{equation}
m_e^{(t-1)} = m^{(t-1)} E_m 
\end{equation}
\subsubsection{Intra-visit Set Encoder} Unlike the previous works \cite{MICRON, safedrug}, which use the code embedding representation of the medical events as the patient representation. We employ the code-level embedding as the input of the set encoder to learn the code-level relationship and then ingrate the code-level information into the visit-level representation. 
Inspired by the Set-Transformer\cite{set_transformer}, we utilize inducing point methods to compress medical code representations into a more compact space for modeling the impact of medical events.
The set encoder contained two \textit{Induced Set Attention Block} (ISAB). In ISAB, along with the set $X \in \mathbb{R}^{m \times d}$, define a new trainable parameters vector $I \in \mathbb{R}^{n \times d}$, called inducing points. The ISAB has the two major sub-layers: \textit{Multi-Head Attention} (MHA) and \textit{row-wise FeedForward layer} (rFF), the functions are defined as:
\begin{equation}
    MHA(Q,K,V) = [head_1, head_2, \cdots, head_h] 
\end{equation}
\begin{equation}
    head_i = Att(QW_i^Q, KW_i^K, VW_i^V) 
\end{equation}
\begin{equation}
    Att(Q,K,V) = Softmax(\frac{QK^{\top}}{\sqrt{s}})V 
\end{equation}
\begin{equation}
    rFF(X) = Relu(XW_{rFF} + b_{rFF})
\end{equation}
where $Q \in \mathbb{R}^{n_q \times d}, K \in \mathbb{R}^{n_k \times d}, V \in \mathbb{R}^{n_v \times d}$ are the inputs of attention $Att(\cdot)$, $W_i^Q \in \mathbb{R}^{d \times d_q}, W_i^K \in \mathbb{R}^{d \times d_k}, W_i^V \in \mathbb{R}^{d \times d_v}$, and $d_q=d_k=d_v = d/h$. $W_{rFF} \in \mathbb{R}^{d \times d}$ and $b_{rFF} \in \mathbb{R}^{d}$ are learnable parameters. The $[\cdot]$ means the concatenate operation. The ISAB is defined as:
\begin{equation}
    ISAB(X) = LN(H+rFF(H))
\end{equation}
\begin{equation}
    H = LN(X+MHA(X,Y,Y))
\end{equation}
\begin{equation}
    Y = LN(Z+rFF(Z))
\end{equation}
\begin{equation}
    Z = LN(I+MHA(I,X,X))
\end{equation}
where $LN$ is layer normalization operation. The set-encoder is defined as:
\begin{equation}
SE_{*}(X) = ISAB(ISAB(X))
\end{equation}
where $* \in \{d,p,m \}$.

Given the code-level embedding representation, the output of the diagnosis set encoder is formulated as follows:
\begin{equation}
    S_d^{(t)} = SE_d(d_e^{(t)}))
\end{equation}

Similar to the diagnosis set encoder, the output of the procedure set encoder and medication set encoder are formulated as $S_p^{(t)} = SE_p(p_e^{(t)}), S_m^{(t-1)} = SE_m(m_e^{(t-1)})$. After obtaining the code-level set representation of the three medical events, we combine them to visit-level representation $V^{(t)}$ as the health status of the patient in the current visit. The visit-level representation is defined as:
\begin{equation}
    V^{(t)} = [V_d^{(t)}, V_p^{(t)}, V_m^{(t-1)}]
\end{equation}
where the $V_d^{(t)}, V_p^{(t)}, V_m^{(t-1)}$ is the summation of code-level representation, and $[\cdot]$ is the concatenate operation.

\subsubsection{Inter-visit Longitudinal Encoder} Previous works usually employ Recurrent Neural Networks (RNN) to model the dynamic patient history for learning longitudinal representations of patients. As the success of the attention mechanism in sequence task \cite{2017attention, Bert, GPT}, it will be helpful to combine the attention mechanism and RNN pattern. Inspired by the Block-Recurrent Transformer (BRT) \cite{BRT}, which applies a transformer layer in a recurrent fashion along the sequence input. Differing from the basic BRT, we have followed the GPT \cite{GPT}, added the masked vector to prevent information leaks while modeling patient longitudinal visit records, and named Recurrent Attention Block (RAB). The RAB mainly includes the update stream between the hidden state vector and the visit-level representation. The hidden state vector carries the patient temporal information, and the visit-level representation updates the information based on the historical state representation. For the state vector, the update function is formulated as follows:
\begin{equation}
    C_{t+1} = g_2(MLP(g_1(C_t^{'}, C_t)), g_1(C_t^{'}, C_t))
\end{equation}
\begin{equation}
    g_*(X,Y) = X \odot f + z \odot i
\end{equation}
\begin{equation}
    f = \sigma(W_fY + b_f + 1)
\end{equation}
\begin{equation}
    i = \sigma(W_iY + b_i - 1)
\end{equation}
\begin{equation}
    z = tanh(W_zY + b_z)
\end{equation}
where $MLP$ is multi-layer perceptron, $\odot$ is the Hadamard product, $W_f \in \mathbb{R}^{n_f \times d_f}, W_i \in \mathbb{R}^{n_i \times d_i}, W_z \in \mathbb{R}^{n_z \times d_z}$ are trainable weight matrices, and $b_f \in \mathbb{R}^{d_f}, b_i \in \mathbb{R}^{d_i}, b_z \in \mathbb{R}^{d_z}$ are trainable bias vectors. The $g_* \in \{g_1, g_2\}$ is the gate mechanism. $C_t^{'}$ is the combination of masked self-attention on the current hidden state $C_t$ and the masked cross-attention with the visit-level representation $V^{(t)}$,
\begin{equation}
    C_t^{'} = W_c^{'}([Att(C_t, C_t, C_t), Att(C_t, V^{(t)}, V^{(t)})]) + b_c^{'}
\end{equation}
where $W_c^{'} \in \mathbb{R}^{n_c^{'} \times d_c^{'}}$ and $b_c^{'} \in \mathbb{R}^{d_c^{'}}$ are learnable parameters. 

The update stream of visit-level representation selects the longitudinal information from the hidden state and visit-level information from the current visit and is defined as:
\begin{equation}
    \hat{V}^{(t)} = MLP(V^{(t)^{'}} + V^{(t)}) + (V^{(t)^{'}} + V^{(t)})
\end{equation}
where $MLP$ is a multi-layer perceptron. $V^{(t)^{'}}$ is the concatenate of visit-level representation masked self-attention and masked cross-attention with the current hidden state, where a central feature is to delegate a considerable portion of the information update responsibility to the process for generating attention weights. The formulation is:
\begin{equation}
    V^{(t)^{'}} = W_v^{'}([Att(V^{(t)},V^{(t)},V^{(t)}), Att(V^{(t)}, C_t, C_t)]) + b_v^{'}
\end{equation}
where $W_v^{'} \in \mathbb{R}^{n_v^{'} \times d_v^{'}}$ and $b_v^{'} \in \mathbb{R}^{d_v^{'}}$ are trainable parameters.

\subsubsection{Adaptive Curriculum Learning module} This module includes the prediction layer and the adaptive curriculum manager. 
After obtaining the updated patient-level representation $\hat{V}^{(t)}$, the final medication representation is generated through an output layer, which is defined as:
\begin{equation}
    \hat{y}^{(t)} = \sigma(W_o \hat{V}^{(t)} + b_o)
\end{equation}
where $\sigma$ is sigmoid function, and $W_o \in \mathbb{R}^{n_v^{'} \times |\mathcal{M}|}$, $b_o \in \mathbb{R}^{|\mathcal{M}|}$ are learnable parameters. 
\begin{itemize}
    \item \textbf{Supervised Multi-label Classification Loss}. The recommendation of medication combinations can be treated as a multi-label prediction task. We use the binary cross entropy loss $l_{bce}$ as the multi-label task loss function, and $l_{bce}$ is defined as:
    \begin{equation}
        \mathcal{L}_{bce} = -\sum_{t}^{v_j}\sum_{i}^{|\mathcal{M}|} m_i^{(t)}log(\hat{y}_i^{(t)}) + (1 - m_i^{(t)})log(1 - \hat{y}_i^{(t)})
    \end{equation}
    where $m_i^{(t)}$ and $\hat{y}_i^{(t)}$ means the medical code at $i-$th coordinate at $t-$th visit.
    \item \textbf{Drug-Drug Interaction Loss}. The DDI loss is designed to control the DDI rate of generated medication combinations. Following the previous work~\cite{safedrug}, it is formulated as:
    \begin{equation}
        \mathcal{L}_{ddi} = -\sum_{t}^{v_j}\sum_{i,j}^{|\mathcal{M}|}(A_d \odot (\hat{y}^{(t)}{}^{\top}\hat{y}^{(t)}))
    \end{equation}
    where $\odot$ is the Hadamard product.
    \item \textbf{Combined Loss Functions}. During the training, we noticed that the accuracy and the DDI rate often increase together, mainly due to the drug-drug interaction in real-world clinical scenarios. It is important to balance the multi-label classification loss and the DDI loss. Finally, we use a penalty weight $\alpha$ over the DDI loss for training. The final loss function is defined as:
    \begin{equation}
        \mathcal{L} = \mathcal{L}_{bce} + \alpha \mathcal{L}_{ddi}
    \end{equation}
    where $\alpha$ is a pre-defined hyperparameter. By presetting different $\alpha$, our SHAPE model could meet a different level of DDI requirements (the details of selecting the $\alpha$ are shown in the DISCUSSION section).
    \item \textbf{Adaptive Curriculum Manager}. As shown in Figure 2 (a), although the medication combinations of most long visit records have been recommended before and are easy to predict, the short one lacking historical medication information is the most frequent situation in real-life clinical scenarios, which may be hard to predict accurately. 
    To address this issue, we propose an adaptive curriculum manager to adaptively assign the complex coefficient of each patient and adopt the curriculum learning framework to train our SHAPE model. Specifically, we combine the visit length of the patient into the training schema, where we calculate $\frac{I+l_t}{I_{max}}$ (i.e., Eq. (28)) to adjust the learning rate at the Adam\cite{adam} optimizer.
    Intuitively, when assigning a lower learning rate to shorter patient visit lengths, the model is guided to learn more complex parameter patterns for those shorter visit records.
    The adaptive curriculum manager is defined as:
    \begin{equation}
        \theta_t = \theta_{t-1} - \frac{\hat{\gamma} {\mu}_t}{\sqrt{{\eta}_t} + \epsilon}
    \end{equation}
    \begin{equation}
        \hat{\gamma} = \gamma(1- \frac{I+l_t}{I_{max}})
    \end{equation}
    \begin{equation}
        {\mu}_t = \frac{\beta_1 \mu_{t-1}+(1-\beta_1)g_t}{1-\beta_1}
    \end{equation}
    \begin{equation}
        {\eta}_t = \frac{\beta_2 \eta_{t-1}+(1-\beta_2)g_t^2}{1-\beta_2}
    \end{equation}
    \begin{equation}
        g_t = \nabla_{\theta}f_t(\theta_{t-1})
    \end{equation}
     where $\epsilon$ is a constant added to the denominator to improve numerical stability, $\gamma$ is the learning rate, $I$ is the current training iteration number, $l_t$ is the current visit length, $I_{max}$ is the pre-defined maximum iteration number, and $\mu_t, \eta_t$ is the parameter of the first moment and the second moment of Adam, $\beta_1, \beta_2$ is the coefficient of the moment, the $f(\theta)$ is the objective function, and $\theta$ are parameters waiting to update, $\nabla(\cdot)$ is the derivative operation. The adaptive curriculum manager is banded with the parameter update. Eq. (28) is the critical step of the optimizer of the objective. We use the current iteration and the current patient visit length to select the learning difficulty automatically. 
\end{itemize}
\subsection{Inference}
The SHAPE is trained end-to-end, and in the inference phase, the safe drug combination recommendation is generated from the sigmoid output $\hat{y}^{(t)}$, where we fix the threshold value as 0.5 to predict the label set. Then, the final predicted medication combinations correspond to the following:
\begin{equation}
    \hat{Y}^{(t)} = \{\hat{y}_i^{(t)}| \hat{y}_i^{(t)} > 0.5, 1\leq i \leq |\mathcal{M}| \}
\end{equation}

\section{Experiments}
In this section, we introduce the experiment details and conduct evaluation experiments to demonstrate the effectiveness of our SHAPE model\footnote{https://github.com/sherry6247/SHAPE.git}. 

\subsection{Dataset}
We use the EHR data from the Medical Information Mart for Intensive Care (MIMIC-III)\footnote{https://mimic.physionet.org/}. It contains 46,520 patients and 58,976 hospital admissions from 2001 to 2012. We conduct experiments on a benchmark released by COGNet \cite{COGNet}, which is based on the MIMIC-III dataset for a fair comparison. Following the COGNet, we selected Top-40 severity DDI types from TWOSIDES\cite{tatonetti2012data}, and we converted the drug code into ATC Third Level codes\footnote{https://www.whocc.no/atc/structure\_and\_principles/} to align with the DDI graph nodes. Finally, we followed the setting of COGNet and divided the dataset into training, validation, and test sets by the ratio of $4:1:1$. The statistics of the post-processed data are reported in Table 1. 

\begin{table}[htbp]
    \caption{Data Statistics}
    \begin{center}
    \begin{tabular}{c|c}
    \hline
    \hline
        Item & Size \\
        \hline
        \# of visits/ \# of patients & 14,995 / 6,350 \\
        diag. / proc. / med. set size & 1,958 / 1,430 / 131 \\
        avg. / max. \# of visits & 2.37 / 29 \\
        avg. / max. \# of diagnoses per visit & 10.51 / 128 \\
        avg. / max. \# of procedure per visit & 3.84 / 50 \\
        avg. / max. \# of medication per visit & 11.44 / 65 \\
        total \# of DDI pairs & 448 \\
    \hline
    \hline
    \end{tabular}
    \label{tab:my_label}
    \end{center}
\end{table}

\subsection{Metrics}
We use three efficacy metrics: Jaccard, F1, and Precision-Recall Area Under Curve (PRAUC) combinations to evaluate the recommendation efficacy. Additionally, we also showed the DDI rate, and the number of predicted medications following the previous works~\cite{safedrug, COGNet}.

The Jaccard for the patient is calculated as below:
\begin{equation}
    Jaccard = \frac{1}{T}\sum_{t=1}^{T} \frac{ |M^{t} \cap \hat{Y}^{(t)}| }{ |M^{t} \cup \hat{Y}^{(t)}| }
\end{equation}
where the $M^{(t)}$ is the ground-truth medication set sequence at $t-$th visit and the $\hat{Y}^{(t)}$ is the predicted medication combinations.

The F1 of the patient is calculated as follows:
\begin{equation}
    F1 = \frac{1}{T}\sum_{t=1}^{T} 2\times \frac{P_t*R_t}{P_t+R_t}
\end{equation}
\begin{equation}
    P_i = \frac{ |M^{i} \cap \hat{Y}^{(i)}| }{|\hat{Y}^{(i)}|}
\end{equation}
\begin{equation}
    R_i = \frac{ |M^{i} \cap \hat{Y}^{(i)}| }{|M^{i}|}
\end{equation}

The PRAUC is calculated with the ground truth code's predicted probability of each medication code.
\begin{equation}
    PRAUC = \frac{1}{T}\sum_{t=1}^{T}\sum_{k=1}^{|\mathcal{M}|}P(k)_t (R(k)_{t} - R(k-1)_{t})
\end{equation}
where $P(k)_t, R(k)_{t}$ are the precision and recall at the cut-off $k-$th threshold in the ordered retrieval list.

DDI rate aims to measure the interaction between the recommended medication combinations, which is calculated as follows:
\begin{equation}
    DDI = \frac{1}{T}\sum_{t=1}^{T} \frac{\sum_{i=1}^{|\hat{Y}^{(t)}|} \sum_{j=i+1}^{|\hat{Y}^{(t)}|} \mathbf{1} \{ A_d[\hat{Y}_i^{(t)}, \hat{Y}_j^{(t)}]=1 \} } { \sum_{i=1}^{|\hat{Y}^{(t)}|} \sum_{j=i+1}^{|\hat{Y}^{(t)}|}  1 }
\end{equation}
where $A_d$ is the known knowledge of the DDI matrix. $\hat{Y}_i^{(t)}$ denoted the $i-$th recommended medication and $\mathbf{1}\{\cdot\}$ means to return 1 when the $\{\cdot\}$ is true, otherwise, return 0.

\begin{table*}[htbp]
    \centering
    \caption{Performance Comparison on the MIMIC-III dataset, the best results are highlighted bold}
    \begin{tabular}{c|c|c|c|c|c}
    \hline
    \hline
          Model & Jaccard & F1 & PRAUC & DDI & Avg.\# of Drugs \\
          \hline
          LR & 0.4865 $\pm$ 0.0021 & 0.6434 $\pm$ 0.0019 & 0.7509 $\pm$ 0.0018 & 0.0829 $\pm$ 0.0009 & 16.1773 $\pm$ 0.0942 \\
          \hline
          LEAP(2017) & 0.4521 $\pm$ 0.0024 & 0.6138 $\pm$ 0.0026 & 0.6548 $\pm$ 0.0033 & 0.0731 $\pm$ 0.0008 & 18.7138 $\pm$ 0.0666 \\
          4SDrug(2022) & 0.4646 $\pm$ 0.0012 & 0.6263 $\pm$ 0.0012 & 0.7604$\pm$ 0.0016 & \textbf{0.0540 $\pm$ 0.0004} & 14.6389 $\pm$ 0.0710 \\
          \hline
          RETAIN(2016) & 0.4887 $\pm$ 0.0028 & 0.6481 $\pm$ 0.0027 & 0.7556 $\pm$ 0.0033 & 0.0835 $\pm$ 0.0020 & 20.4051 $\pm$ 0.2832 \\
          DMNC(2018) & 0.4864 $\pm$ 0.0025 & 0.6529 $\pm$ 0.0030 & 0.7580 $\pm$ 0.0039 & 0.0842 $\pm$ 0.0011 & 20.0000 $\pm$ 0.0000 \\
          GAMNet(2019) & 0.5067 $\pm$ 0.0025 & 0.6626 $\pm$ 0.0025 & 0.7631 $\pm$ 0.0030 & 0.0864 $\pm$ 0.0006 & 27.2145 $\pm$ 0.1141 \\
          MICRON(2021) & 0.5100 $\pm$ 0.0033 & 0.6654 $\pm$ 0.0031 & 0.7687 $\pm$ 0.0026 & 0.0641 $\pm$ 0.0007 & 17.9267 $\pm$ 0.2172 \\
          SafeDrug(2021) & 0.5213 $\pm$ 0.0030 & 0.6768 $\pm$ 0.0027 & 0.7647 $\pm$ 0.0025 & 0.0589 $\pm$ 0.0005 & 19.9178 $\pm$ 0.1604 \\
          COGNet(2022) & 0.5336 $\pm$ 0.0011 & 0.6869 $\pm$ 0.0010 & 0.7739 $\pm$ 0.0009 & 0.0852 $\pm$ 0.0005 & 28.0903 $\pm$ 0.0950 \\
          \textbf{SHAPE} & \textbf{0.5513 $\pm$ 0.0009} & \textbf{0.7017 $\pm$ 0.0008} & \textbf{0.7906 $\pm$ 0.0009 } & 0.0677 $\pm$ 0.0003 & 20.9949 $\pm$ 0.1189 \\
    \hline
    \hline
    \end{tabular}
\end{table*}

\begin{figure*}[htbp]
	\centering
	\includegraphics[width=140mm]{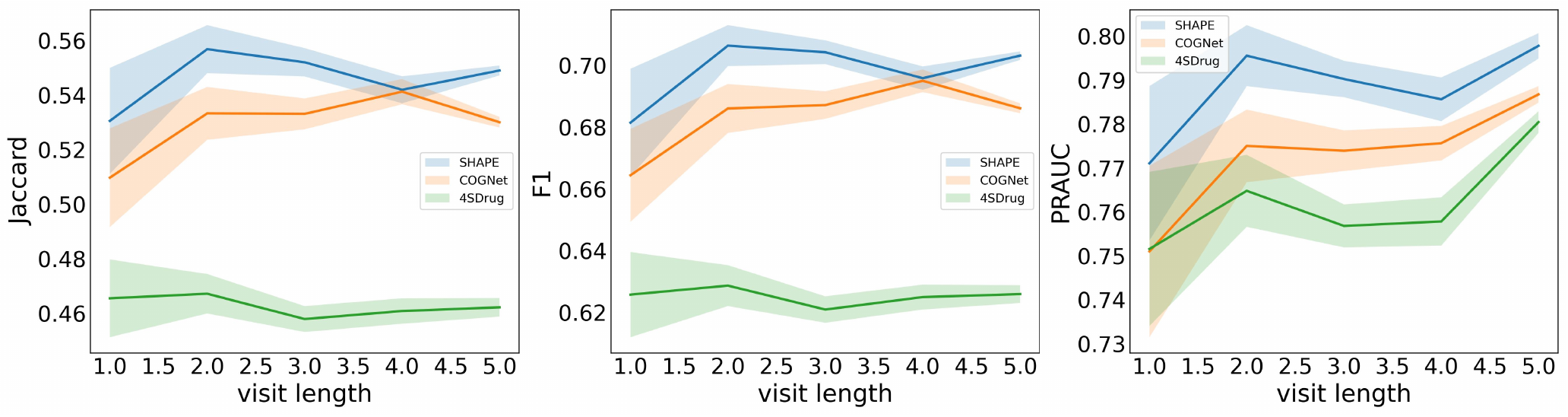}
	\caption{The performance of different visit lengths with the various models.}
\end{figure*}

\subsection{Baseline}
We compare the SHAPE model with the following methods from different perspectives: \textit{conventional machine learning method}, such as Logistic Regression(LR). \textit{Instance-based methods}: LEAP \cite{leap}, 4SDrug \cite{4sdrug}. \textit{Longitudinal-based methods}: RETAIN\cite{retain}, DMNC \cite{dmnc}, GAMENet \cite{gamenet}, MICRON \cite{MICRON}, SafeDrug \cite{safedrug}, COGNet \cite{COGNet}. Specifically, {LEAP}\cite{leap} uses an attention mechanism to encode the diagnosis sequence step by step. {4SDrug} \cite{4sdrug} designs an attention-based method to augment the symptom representation and leverages the DDI graph to generate the current drug sequence. {RETAIN} \cite{retain} employs the attention gate mechanism to model the patient longitudinal information. {DMNC} \cite{dmnc} proposes a memory network to capture more interaction in the patient EHR record. {GAMNet} \cite{gamenet} combines the RNN and graph neural network to recommend medication combinations. {MICRON} \cite{MICRON} leverages a residual-based network to update the patient representation according to the new feature change. {SafeDrug} \cite{safedrug} utilizes drugs' molecule structures in the medication recommendation. {COGNet} \cite{COGNet} proposes a conditional generation model to copy or predict drugs according to the patient representation.


\subsection{Parameter Setting}
Here, we list the implementation details of SHAPE. We set the hidden dimension as 128 and use the Adam optimizer\cite{adam} with an initial learning rate $1 \times 10^{-3}$ for 50 epochs. We fixed the random seed as 2023 to ensure the reproducibility of the model. Our model is implemented by Pytorch 1.7.1 based on Python 3.8.13 and training on two GeForce RTX 3090 GPUs, and an early-stopping mechanism was utilized. For a fair comparison, in the testing stage, we follow the previous work CONGNet \cite{COGNet}, which random sample 80\% data from test data for a round of evaluation. We repeat this process $10$ times and calculate the mean and standard deviation as the final result we reported. 

\begin{table*}[htbp]
    \centering
    \caption{Ablation study for different SHAPE modules on MIMIC-III dataset.}
    \begin{tabular}{c|c|c|c|c|c|c|c|c}
    \hline
    \hline
    Model & ISE & ILE & ACLM & DDI loss & Jaccard & F1 & PRAUC & DDI rate \\
    \hline
    SHAPE & $\checkmark$ & $\checkmark$ & $\checkmark$ & $\checkmark$ & 0.5513 $\pm$ 0.0009 & 0.7017 $\pm$ 0.0008 & 0.7906 $\pm$ 0.0009 & 0.0677 $\pm$ 0.0003  \\
    SHAPE$_{w SA}$ &  & $\checkmark$ & $\checkmark$ & $\checkmark$ & 0.5404 $\pm$ 0.0008 & 0.6922 $\pm$ 0.0013 & 0.7845 $\pm$ 0.0011 & 0.0681 $\pm$ 0.0007 \\
    SHAPE$_{w/o ISE}$ &  & $\checkmark$ & $\checkmark$ & $\checkmark$ & 0.5280 $\pm$ 0.0011 & 0.6828 $\pm$ 0.0010 & 0.7739 $\pm$ 0.0009 & 0.0716 $\pm$ 0.0005 \\
    SHAPE$_{w/o ILE}$ & $\checkmark$ & & $\checkmark$ & $\checkmark$ & 0.5243 $\pm$ 0.0016 & 0.6793 $\pm$ 0.0014 & 0.7718 $\pm$ 0.0018 & 0.0699 $\pm$ 0.0003 \\
    SHAPE$_{w/o ACLM}$ & $\checkmark$ & $\checkmark$ & & $\checkmark$ & 0.5314 $\pm$ 0.0020 & 0.6856 $\pm$ 0.0018 & 0.7768 $\pm$ 0.0020 & 0.0660 $\pm$ 0.0004 \\
    SHAPE$_{w/o DDI loss}$ & $\checkmark$ & $\checkmark$ & $\checkmark$ & & 0.5483 $\pm$ 0.0016 & 0.6989 $\pm$ 0.0014 & 0.7880 $\pm$ 0.0012 & 0.0857 $\pm$ 0.0005  \\
    \hline
    SHAPE & $\checkmark$ & $\checkmark$ & $\checkmark$ & $\alpha_{ddi}$=0.1 & 0.5411 $\pm$ 0.0014 & 0.6934 $\pm$ 0.0014 & 0.7848 $\pm$ 0.0012 & 0.0559 $\pm$ 0.0003 \\
    SHAPE & $\checkmark$ & $\checkmark$ & $\checkmark$ & $\alpha_{ddi}$=0.09 &
    0.5421 $\pm$ 0.0012 & 0.6945 $\pm$ 0.0011 & 0.7843 $\pm$ 0.0012 & 0.0571 $\pm$ 0.0004 \\
    SHAPE & $\checkmark$ & $\checkmark$ & $\checkmark$ & $\alpha_{ddi}$=0.08 &
    0.5451 $\pm$ 0.0017 & 0.6968 $\pm$ 0.0015 & 0.7850 $\pm$ 0.0015 & 0.0601 $\pm$ 0.0003 \\
    SHAPE & $\checkmark$ & $\checkmark$ & $\checkmark$ & $\alpha_{ddi}$=0.05 &
    \underline{0.5513 $\pm$ 0.0009} & \underline{0.7017 $\pm$ 0.0008} & \underline{0.7906 $\pm$ 0.0009} & \underline{0.0677 $\pm$ 0.0003} \\
    SHAPE & $\checkmark$ & $\checkmark$ & $\checkmark$ & $\alpha_{ddi}$=0.02 & 0.5507 $\pm$ 0.0012 & 0.7015 $\pm$ 0.0010 & 0.7919 $\pm$ 0.0010 & 0.0787 $\pm$ 0.0004 \\
    SHAPE & $\checkmark$ & $\checkmark$ & $\checkmark$ & $\alpha_{ddi}$=0.01 & 0.5496 $\pm$ 0.0017 & 0.7009 $\pm$ 0.0014 & 0.7893 $\pm$ 0.0019 & 0.0844 $\pm$ 0.0005 \\
    \hline
    \hline
    \end{tabular}
\end{table*}
\begin{table*}[htbp]
    \centering
    \caption{Example recommended medications for a given patient health condition on MIMIC-III.}
    \begin{tabular}{c|l|c|c|c} 
        \hline
        \hline
        \textbf{Case 1} & \multicolumn{4}{l}{\textbf{Code}}  \\
        \hline
        Diagnosis: & \multicolumn{4}{l}{03849,5770,5849,5761,1623,5859,496,99592,57450,7904,28800,V1582,40390,4439,2720,28522,1991}  \\
        Procedure:  & \multicolumn{4}{l}{5185,5188,5187,5293} \\
        Ground truth: & \multicolumn{4}{l}{D06A,J01C,N05A,A04A,J01M,A12A,A07A,N02A,A12B,A01A,B05C,N02B,B01A,A12C,A02B,A06A (16 codes)} \\
        \hline
        Model & Predicted codes & hit & missed & error \\
        \hline
        4SDrug &  \makecell[l]{N02B,A01A,A02B,A06A,B05C,A12A,A12C,A07A,C03C,A12B,N02A,J01M,B01A,J01D,A04A,R03A,R01A} & 13 & 3 & 4\\
        \hline
        COGNet & \makecell[l]{N02B,A01A,A02B,A06A,B05C,A12A,A12C,C01C,A07A,C07A,A12B,N02A,J01M,B01A,R03A,R01A,J01C} & 13 & 3 & 4 \\
        \hline
        SHAPE & \makecell[l]{N02B,A01A,A02B,A06A,B05C,A12A,A12C,A07A,C07A,C03C,A12B,N02A,J01M,B01A,A04A,J01C} & 14 & 2 & 2\\
        \hline
        \hline
        \textbf{Case 2} & \multicolumn{4}{l}{\textbf{Code}} \\
        \hline
        Diagnosis: & \multicolumn{4}{l}{03842,78552,V427,99592,4019,25000,V1007}  \\
        Procedure:  & \multicolumn{4}{l}{3893,03311} \\
        Ground truth: & \multicolumn{4}{l}{\makecell[l]{J05A,D11A,D01A,J01E,R05C,A07E,C01C,N05B,J01D,N06A,A12A,A07A,N02A,A12B,A01A,B05C,N02B,B01A,\\A12C,A02B,A06A (21 codes)}} \\
        \hline
        Model & Predicted codes & hit & missed & error \\
        \hline
        4SDrug & \makecell[l]{N02B,A01A,A02B,A06A,B05C,A12C,C01C,A07A,C07A,C03C,N02A,B01A,J01D,D11A,A07E,J05A,J01E,\\L04A} & 15 & 6 & 3 \\
        \hline
        COGNet & \makecell[l]{N02B,A01A,A02B,A06A,B05C,A12A,A12C,C01C,A07A,C07A,A12B,C02D,N06A,B01A,D01A,N05A,\\D11A,A04A,A07E,J05A,J01E,J01C,C02C} & 17 & 4 & 6 \\
        \hline
        SHAPE & \makecell[l]{N02B,A01A,A02B,A06A,B05C,A12A,A12C,C01C,A07A,A12B,N02A,N06A,B01A,D01A,J01D,D11A,\\A07E,J05A,J01E,J01C,L04A} & 19 & 2 & 2 \\
        \hline
        \hline
    \end{tabular}
\end{table*}

\subsection{Result Analysis}
As shown in Table 2, our proposed model SHAPE outperforms all baselines with the higher Jaccard, F1, and AUPRC and increased by nearly 2\% compared to the previous best model. The conventional LR and the Instance-based methods are poor as they only consider the patient's health condition at the current visit. The performance of RETAIN and DMNC are comparable because both use the RNN architecture to capture the longitudinal information. The GAMENet introduced an additional DDI graph and fused it with the EHR co-occurrence graph, resulting in further performance improvement. SafeDrug leverages the drugs' molecule structures to improve the performance of medication recommendations. Unlike most longitudinal algorithms, which focus on the historical record, the MICRON proposed using the residual network to capture changes in medications. The COGNet proposes the copy or prediction mechanism to generate the medication sequence since the statistics show that most medication codes have been recommended in historical EHR records. However, it fails to consider the short visit, which may not be enough historical reference, especially for the newly and secondly admission patients. 

Compared with the baseline methods, our SHAPE model achieves state-of-the-art performance. On the one hand, it designed an intra-visit set encoder to automatically collect the most informative medical events of each patient. On the other hand, we develop an inter-visit longitudinal encoder to capture the longitudinal pattern, which inherits the merit of RNN and the attention mechanism. Besides, our adaptive curriculum manager assigns the difficulty of each sample base on the visit length accordingly. Hence, our SHAPE performance is better than the other methods. 

We also noticed in Table 2 that the 4SDrug achieves the lowest and most charming DDI rate of predicted medication combinations. However, combined with the results shown in Figure 4, the 4SDrug achieves the lowest DDI probably because the predicted medication code counts are less than other methods since we have observed that the DDI rate increase with the number of predicted medications. This lower DDI rate phenomenon also appears in the MICRON model since there are few predicted medications.

Furthermore, we noticed that the MIMIC-III dataset has an average DDI rate of 0.0875 itself, which means there is a large number of DDI phenomena in real-world practice. Based on this fact, our SHAPE also achieves a lower DDI rate and higher accuracy of medication recommendations, indicating  the effectiveness of our proposed method.

To further validate that our SHAPE model can better model the short visit and even the new visit problem and recommend medication effectively, we investigate the performance of various visits with different models. As shown in the right picture of Figure 1, there are severe long-tail phenomena in the MIMIC-III dataset, and most patients have less than five times admission records. We take patients' first five visit records in the test set for visualization. We compared SHAPE with the COGNet and 4SDrug since (1) the COGNet achieves the best performance of the existing methods, and (2) the 4SDrug method uses the set-orient method to learn the code-level representation and uses the DDI loss to control the output predicted. As shown in Figure 4, our SHAPE model is superior to the COGNet on the three metrics (i.e., Jaccard, F1, and PRAUC). Especially, our SHAPE achieves higher performance in the short visit length and shows an increasing trend. These results may directly show the power of SHAPE to solve the problem shown in Figure 1, in which the short visit records are the critical samples. The higher accuracy of these samples is helpful for most situations in real-world clinical practices. On the contrary, the 4SDrug is always under the COGNet and SHAPE. The reason may be that the 4SDrug is an instance-based method that ignores temporal longitudinal information.

\begin{figure}[htbp]
	\centering
	\includegraphics[width=70mm]{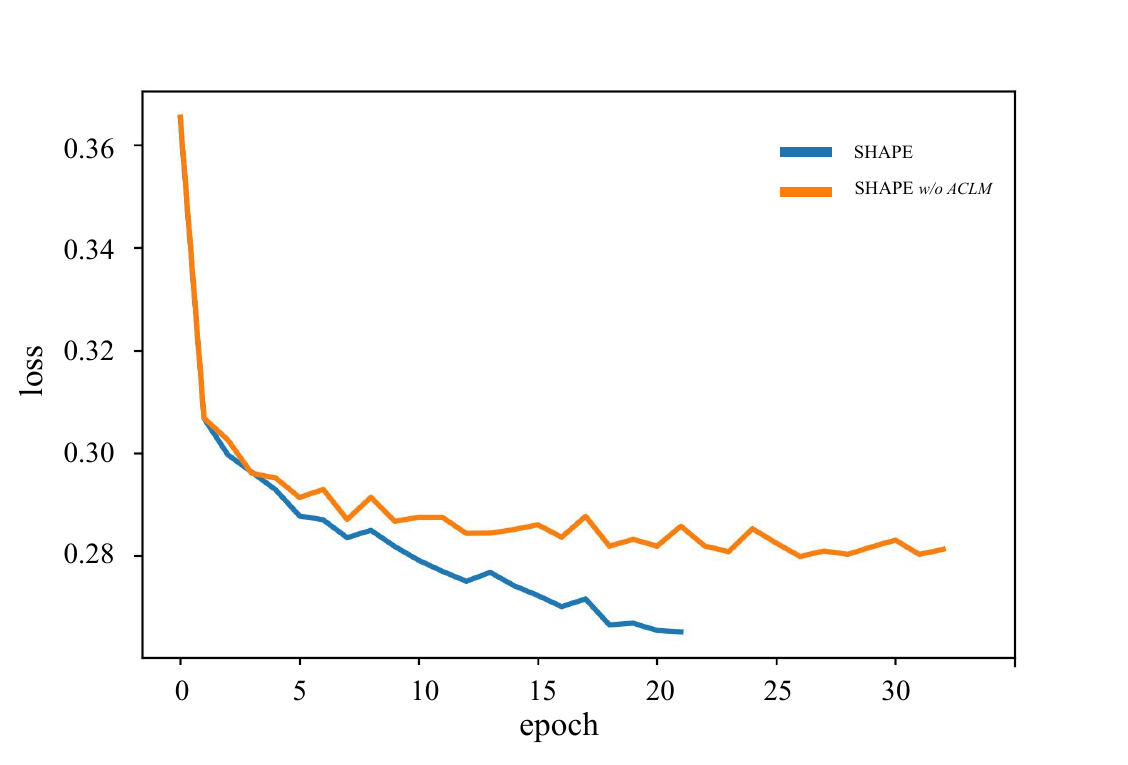}
	\caption{Loss comparison on SHAPE and SHAPE$_{w/o ACLM}$ regarding different numbers of train epochs.}
\end{figure}

\begin{figure*}[htbp]
	\centering
	\includegraphics[width=155mm]{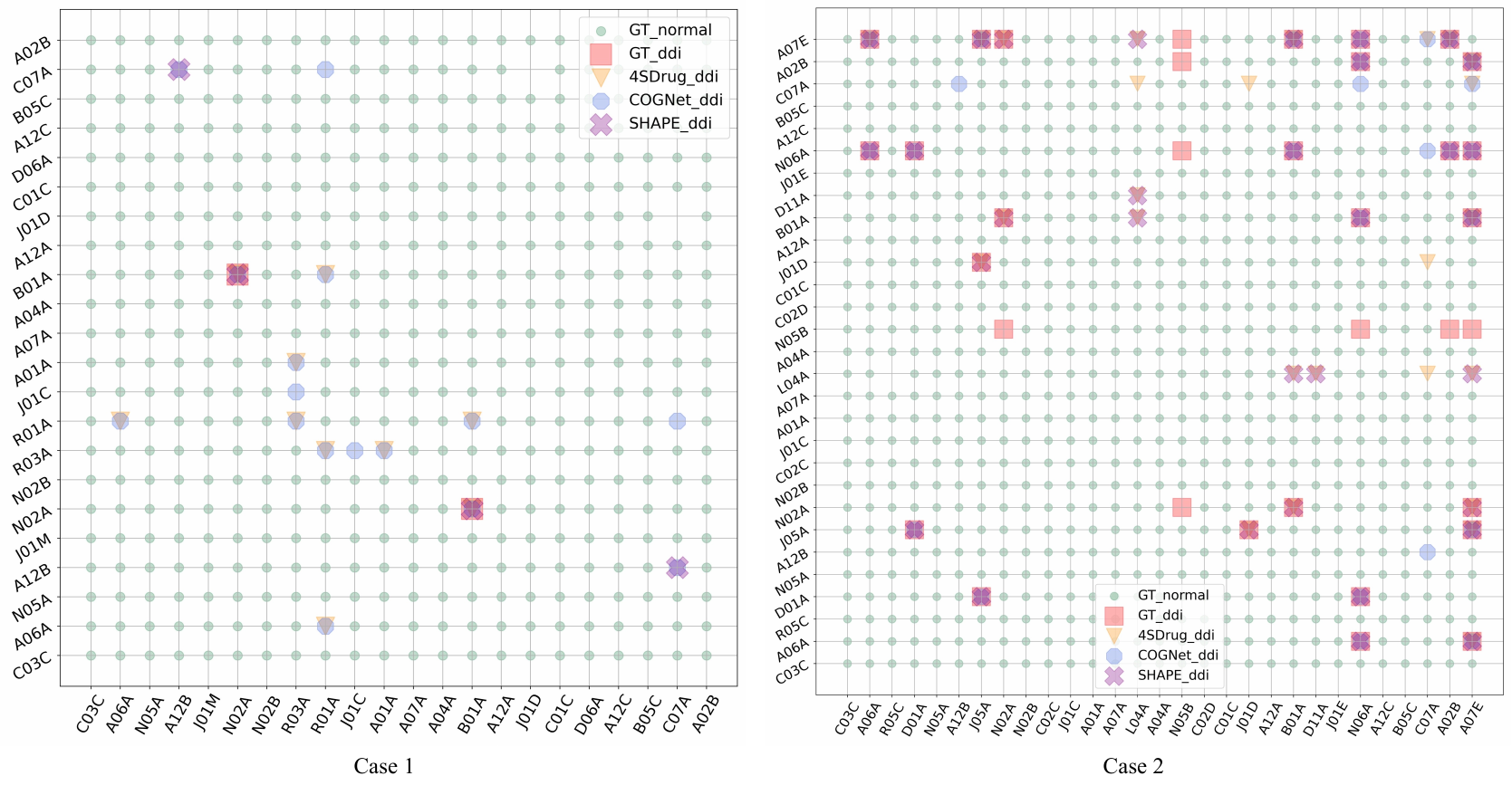}
	\caption{Visualization DDI of the case study. \textit{Case 1} is a new admission patient. \textit{Case 2} is a secondary admission patient. In a chessboard, the red square corresponds to the DDI in the ground truth; the green point corresponds there are not appear DDI in the ground truth; the blue circle corresponds to the DDI in the predicted medications with COGNet, the inverted yellow triangle corresponds to the DDI predicted medications with 4SDrug. The purple cross corresponds to the DDI in the predicted medications with SHAPE. Best viewed in color.}
\end{figure*}

\section{Discussion}
Upon analyzing the results in Table 2, we can conclude that our proposed model SHAPE achieved the best performance compared to the LR and \textit{Instance-base} and \textit{Longitudinal-base} methods. The success of SHAPE is ascribed to the three modules we proposed (i.e., the Intra-visit Set Encoder (ISE), the Inter-visit Longitudinal Encoder (ILE), and Adaptive Curriculum Learning Module (ACLM)), and it achieved a lower DDI rate with our proposed combined loss function. To verify the effectiveness of each module we proposed, we designed the ablation experiments, SHAPE$_{w/o ISE}$: which remove the intra-visit set encoder and summarize the code-level to visit-level representation directly. SHAPE$_{w/o ILE}$: which uses the recurrent neural network to replace the inter-visit longitudinal encoder for learning the longitudinal information. SHAPE$_{w/o ACLM}$: which means removing the step of Eq. (28) and using the basic Adam optimizer to optimize the SHAPE. SHAPE$_{w/o DDI loss}$: which only uses the multi-label classification loss function as the objective to train the model. We also compared the self-attention (SA) to investigate the effectiveness of our proposed compact intra-visit set encoder, SHAPE$_{w SA}$: which replaces the set encode as self-attention.

Table 3 shows the results for the different variants of SHAPE. As expected, when randomly removing the three modules we proposed. The performance brought a significant deterioration to the complete SHAPE model. The results of the DDI rate of SHAPE$_{w/o DDI loss}$ illustrate the effectiveness of the combination loss function. Overall, the SHAPE outperforms all variant models, which means each component is integral to SHAPE. Compared with the SHAPE, the SHAPE$_{w SA}$ drops performance on total metrics, demonstrating that a more compacted encoder is more suitable to model the complex medical event code sequence. 

Moreover, the performance drop of SHAPE$w/o ACLM$ can be observed in Table 3, indicating that it is important to consider the visit length as the guidance to assign the complex coefficient in the model of each patient. To explore the impact of the ACLM module, we conducted experiments to visualize the loss trajectory between SHAPE and SHAPE$w/o ACLM$. As shown in Figure 5, it can be seen that compared to SHAPE$w/o ACLM$, SHAPE has a significant decrease in loss and converges quickly. This demonstrates the vital importance of the ACLM module, as it can automatically assign difficulty coefficients to each sample and learn more suitable parameters for various visit records. 

Furthermore, to achieve a satisfactory trade-off for the DDI rate phenomenon in the medication combinations generated by SHAPE, we explore the hyperparameter $\alpha$ in Eq. (26). The details are also shown in the second half of Table 3, according to the results of Table 3, we can conclude that: (1) the DDI rate of predicted medication combinations is gradually increasing with the decline of $\alpha_{ddi}$. (2) before the $\alpha>0.05$, the performance of other metrics is suppressed, which indicates the DDI rate and the accuracy performance of the predicted medication combination almost linearly decreases with the penalty weight. However, when the $\alpha < 0.05$, the performance of SHAPE fluctuated. Combined with the previously mentioned that the MIMI-III dataset has a 0.0875 DDI rate itself, which means not the lowest DDI rate is the superior optimal selection of clinical practice.

To intuitively demonstrate the advantages of SHAPE over the two baseline models, we analyze some examples to show the predicted results. We choose the short or new visit patients to demonstrate the model effect on harder predicted cases. Due to space constraints, we use the International Classification of Disease (ICD) code to represent the diagnosis and procedure information and the ATC code to represent the medications. As shown in Table 4, \textit{case 1} is a new admission patient, the doctor prescribed ground truth medication based on the diagnosis and procedure information of the patient's current visit. \textit{Case 2} is a secondary admission patient, and we list the second record in \textit{Case 2}. In \textit{Case 2}, the physician combines the current health condition and the patient's historical record to prescribe medication. Overall, the SHAPE performed the best with 14 correct and 19 correct medications in two cases and achieved the lowest miss or error in the two cases. Furthermore, we noticed that in new visit \textit{Case 1}, the instance-based method 4SDrug also achieves comparable performance with COGNet, probably because of the instance-based approach against the single visit problem.

As shown in Figure 6, we visualize the DDI status in two cases of each model, where the symmetric matrix shows the drug-drug relationship of the combination of medications. The point of $GT_{normal}$ means there is no DDI in ground truth medication combinations, and $GT_{ddi}$ means there probably is DDI in the ground truth medication combinations. The empty rows and columns mean these codes do not appear in the ground truth medications. We noticed in \textit{Case 1} our SHAPE only generates two pairs of medication which maybe suffers the drug-drug interaction, on the contrary, the 4SDrug and COGNet generate five pairs (i.e., [A01A, R03A], [A06A, R01A], [N02A, B01A], [B01A, N02A], [B01A, R01A]) and eight pairs (i.e., [A01A, R03A], [A06A, R01A], [C07A, A12B], [C07A, R01A], [A12B, C07A], [N02A, B01A], [B01A, N02A], [B01A, R01A]).  In the DDI of \textit{Case 2}, we find that the DDI phenomenon in real-life scenarios exceeds ten pairs of medications. Our SHAPE simultaneously hits most situations similar to the ground truth medication prescribed by doctors, which hints that SHAPE can provide a safer way to recommend medication combinations.

There are also several limitations of the current study. Firstly, we only used diagnosis and procedure information for the side information to infer the medication and ignored others, such as vital signs and laboratory test records. Secondly, we only evaluate the SHAPE model on a public dataset, which also limits the generalizability of the model.

\section{Conclusion}
In this paper, we proposed a sample adaptive hierarchical medication prediction network, named SHAPE, to better learn the accurate representation of the patient. Concretely, we first present an intra-visit set encoder to capture medical events relationship from the code-level perspective, which is usually ignored in most current works. Then, we developed an inter-visit longitudinal encoder to learn the visit-level longitudinal representation, which inherits the merits between attention and the RNN. Additionally, we designed an adaptive curriculum learning module that references patients' personalities to automatically assign each patient's difficulty for improving the performance of medication recommendations. 
Experiment results on the public benchmark dataset demonstrate that SHAPE outperforms existing medication recommendation algorithms by a large margin. We also investigate the performance of short visits and new visit samples, which shows that the SHAPE can effectively figure out the medication recommendation with the short admission of patients. Further ablation study results also suggest the effectiveness of each module of our proposed SHAPE. 

\bibliographystyle{ieeetr}
\bibliography{ref}

\end{document}